\documentclass[letterpaper]{article} 
\usepackage{aaai24}
\usepackage{times}  
\usepackage{helvet}  
\usepackage{courier}  
\usepackage[hyphens]{url}  
\usepackage{graphicx} 
\usepackage{natbib}  
\usepackage{caption} 
\usepackage{algorithm}
\usepackage{algorithmic}

\usepackage{amsmath} 
\usepackage{bm}
\usepackage{booktabs} 
\usepackage{multirow}
\usepackage{amsthm}

\usepackage[dvipsnames, svgnames, x11names]{xcolor} 
\usepackage{colortbl} 
\usepackage{threeparttable}
\usepackage{array}
\usepackage{microtype}
\usepackage{graphicx}
\usepackage{bm}
\usepackage{algorithm}
\usepackage{algorithmic}
\usepackage{multirow}
\usepackage{amsthm}
\usepackage{amsmath}  
\usepackage{arydshln}

\usepackage{pifont}    
\usepackage{bbding}

\usepackage{graphicx}

\usepackage{newfloat}
\usepackage{listings}
\DeclareCaptionStyle{ruled}{labelfont=normalfont,labelsep=colon,strut=off} 
\floatstyle{ruled}
\newfloat{listing}{tb}{lst}{}
\floatname{listing}{Listing}
 
\title{Diving into Darkness: A Dual-Modulated Framework for High-Fidelity Super-Resolution in Ultra-Dark Environments}
\author{
	Jiaxin Gao\textsuperscript{\rm 1}, 
	Ziyu Yue\textsuperscript{\rm 2}, 
	Yaohua Liu\textsuperscript{\rm 1},
	Sihan Xie\textsuperscript{\rm 1},
	Xin Fan\textsuperscript{\rm 1},
	Risheng Liu\textsuperscript{\rm 1} 
}
\affiliations{
    \textsuperscript{\rm 1} School of Software Technology, Dalian University of Technology, China\\
    \textsuperscript{\rm 2} School of Mathematical Science, Dalian University of Technology, China\\ 

Jiaxinn.gao@outlook.com, 11901015@mail.dlut.edu.cn, liuyaohua\_918@163.com, XSH2018@mail.dlut.edu.cn, xin.fan@dlut.edu.cn, rsliu@dlut.edu.cn\\
}

\usepackage{bibentry}

\begin{document}

\maketitle

\begin{abstract}
Super-resolution tasks oriented to images captured in ultra-dark environments is a practical yet challenging problem that has received little attention.  Due to uneven illumination and low signal-to-noise ratio in dark environments, a multitude of problems such as lack of detail and color distortion may be magnified in the super-resolution process compared to normal-lighting environments.  Consequently, conventional low-light enhancement or super-resolution methods, whether applied individually or in a cascaded manner for such problem, often encounter limitations in recovering luminance, color fidelity, and intricate details. To conquer these issues, this paper proposes a specialized dual-modulated learning framework that, for the first time, attempts to deeply dissect the nature of the low-light super-resolution task. Leveraging natural image color characteristics, we introduce a self-regularized luminance constraint as a prior for addressing uneven lighting. Expanding on this, we develop Illuminance-Semantic Dual Modulation (ISDM) components to enhance feature-level preservation of illumination and color details. Besides, instead of deploying naive up-sampling strategies, we design the Resolution-Sensitive Merging Up-sampler (RSMU) module that brings together different sampling modalities as substrates, effectively mitigating the presence of artifacts and halos.  
Comprehensive experiments showcases the applicability and generalizability of our approach to diverse and challenging ultra-low-light conditions, outperforming state-of-the-art methods with a notable improvement (i.e., $\uparrow$5\%  in PSNR, and $\uparrow$43\% in LPIPS). Especially noteworthy is the 19-fold increase in the RMSE score, underscoring our method's exceptional generalization across different darkness levels.
The code will be available online upon publication of the paper.
\end{abstract}

\section{Introduction}

Image super-resolution is a classical problem in low-level vision. Nevertheless, research towards Low-Light Image Super-Resolution (LLISR) has been relatively neglected, despite its practical importance for applications such as surveillance, automated driving, and medical image analysis.  This study aims to explore the enhancement of low-resolution images captured under ultra-dark conditions to achieve both brightness adjustment and magnification, and ultimately generates high-resolution clear images with normal illumination. 

In comparison to well-lit scenarios, the super-resolution process can amplify a range of issues arising from low-light conditions, such as inadequate details and color distortion due to uneven illumination. Typically, illumination distribution is uneven in dimly lit environments, potentially resulting in pronounced shadows and highlight regions. Hence, handling these lighting variations is essential during super-resolution to ensure increased resolution while retaining illumination details. Moreover, due to shooting in low-light conditions, missing details may result in inaccurate or numerous artifacts and halos in the super-resolution results.
 
\begin{figure}[!t]
	\centering 
	\begin{tabular}{c}
		\includegraphics[width=8.2cm]{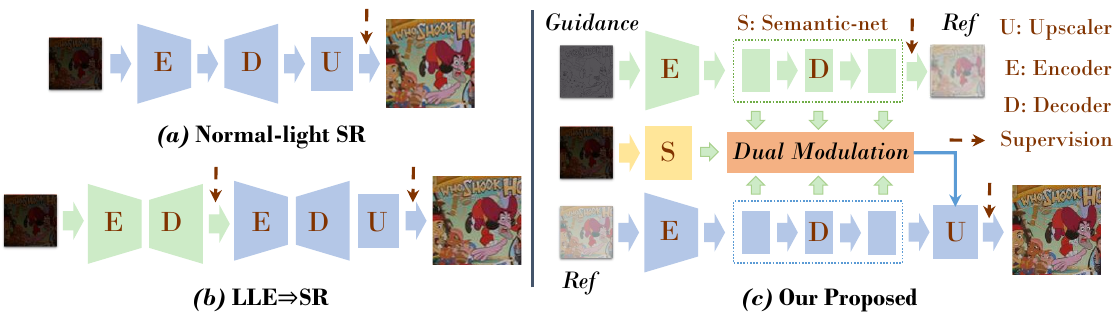}\\
	\end{tabular}%
	\caption{
		Comparative illustration showcasing the fundamental architectures towards LLISR of various solutions.
	} \label{fig:fig1}
	\vspace{-3mm}
\end{figure}
\begin{figure}[!t]
	\centering 
	\begin{tabular}{c}
		\includegraphics[width=8.2cm]{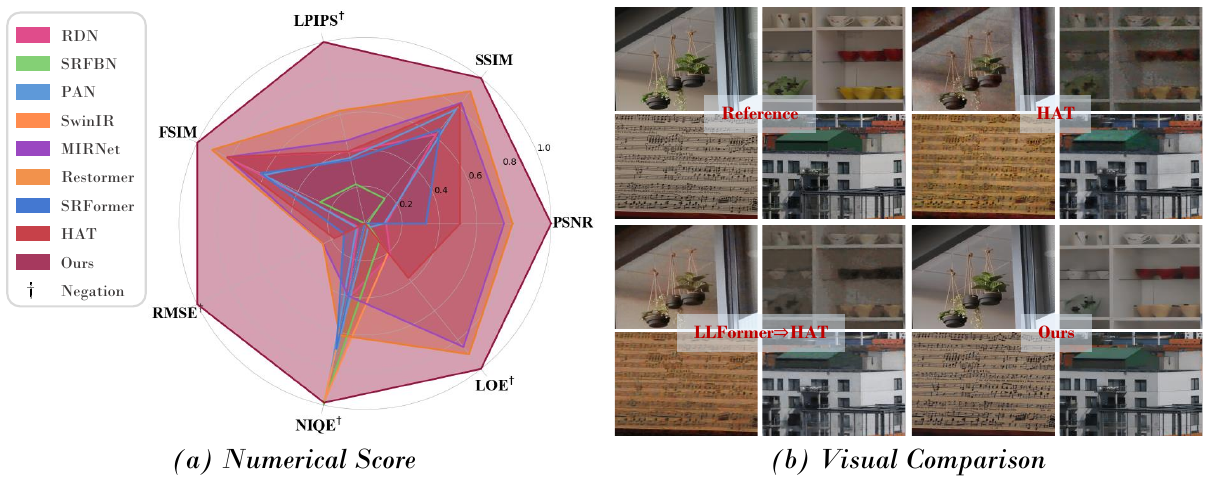}\\
	\end{tabular}%
	\caption{
		Comparisons among different schemes for LLISR in terms of numerical scores \textit{(a)}, as well as visualization results \textit{(b)}. A recent LLE method (i.e., LLFormer~\cite{wang2023ultra}), and a normal-light SR method (HAT~\cite{chen2023hat}), are presented as examples for visualization.  
	} \label{fig:fig2}
	\vspace{-3mm}
\end{figure}

%
Evidently, it has been confirmed by extensive experiments that current methods designed for low-light image enhancement (LLE) and normal-light image super-resolution (SR)  exhibit noticeable limitations in solving the LLISR problem.  Furthermore, simply applying these task-specific models in a cascaded manner (e.g., LLE$\Rightarrow$SR) still fails to yield satisfactory results. The aforementioned findings have been extensively discussed in the literature~\cite{aakerberg2021rellisur}.  \textit{Figure}~\ref{fig:fig1} illustrates the qualitative and quantitative outcomes of various schemes \textit{(a-c)} for handling LLISR tasks. It is important to emphasize that all the methods involved have been retrained on the specific RELLISUR dataset~\cite{aakerberg2021rellisur}. As ilustrated in \textit{Figure}~\ref{fig:fig2}, the normal-light SR technique HAT~\cite{chen2023hat} fails to reconstruct the fine details of low-light input images, resulting in significant artifacts. The LLE method LLFormer~\cite{wang2023ultra} cascaded with HAT exhibits noticeable color shifts and unclear textures.  In comparison, our proposed method shows more natural and realistic colors and structural details. For further comparative analysis results, please refer to the Experimental Section.

To address these issues, this paper proposes a specialized low-light super-resolution paradigm for simultaneously \textbf{B}rightening and \textbf{M}agnifying low-resolution images captured in \textbf{Ultra}-dark scenes, dubbed \textbf{UltraBM}.  
UltraBM utilizes a dual-stream learning framework, which represents a pioneering effort to comprehensively analyze the essence of the LLISR task. During the initial stage, we propose incorporating illumination prior to impose constraints on low-light scenes in an unsupervised manner. Whereas, in the subsequent stage, we introduce a semantic knowledge base, and build Illumination-Semantic Dual Modulation (ISDM) as refinement middleware.  ISDM runs through the entire network layer of decoding for cross-branch interaction, facilitating the far-flung retention of illuminance details and semantic color details, to enable the super-resolution network to focus more on faithful texture details.  
Furthermore, we introduce a Resolution-Sensitive Merging Up-sampler (RSMU) module, a strategic departure from naive up-sampling, effectively alleviating artifacts and halos.
We conducted extensive experiments to validate the effectiveness of the proposed method.
The main contributions can be summarized as follows:
\begin{itemize}
	\item 
	Aiming for the highly-coupled complex task of simultaneously brightening and magnifying images captured in ultra-dark scenes, we introduce a tailored high-fidelity super-resolution framework dubbed UltraBM, marking a 	novel attempt to deeply dissect the nature of the LLISR task and proved feasible solution strategies.
	\item To uphold illuminance details while enhancing resolution, we model low-light scenes by imposing principled priors as initial constraints.  And further we construct a refinement middleware, Illumination-Semantic Dual Modulation (ISDM) that operates in a top-down manner to modulate reflection features, which effectively avoids exacerbating color distortions and emphasizes faithful texture details.
	\item Instead of employing naive up-sampling rules, we assemble different sampling modalities as substrates, thereby architecting a Resolution-Sensitive Merging Up-sampler (RSMU) that adeptly mitigates artifacts and halos, particularly tailored for the refinement of high-fidelity images.
\end{itemize}
We evaluate the applicability and generalizability of the proposed method in diverse extremely ultra-dark scenes, and demonstrate its strengths through comprehensive analysis. 

\section{Related Work}

\subsection{Low-Light Image Enhancement} 

LLE aims to make images hidden in the dark visible.The most widely circulated model is based on the Retinex theory of separating illumination and reflection image layers for reconstruction~\cite{wei2018deep,wang2019underexposed,zhang2019kindling,ren2020lr3m,gao2023learning,liu2021investigating}. In recent years, great progress has also been made in designing models based on convolutional neural networks~\cite{hira2021delta,li2017joint,yang2020fidelity,jiang2021enlightengan,wang2018gladnet}. 
For example, ZeroDCE~\cite{li2021learning} achieved effective enhancement of low-light images by learning reference-free histogram stretching and enhancement functions.
Recently, an unsupervised method SCI was proposed in~\cite{ma2022toward} to use a basic illuminance learning module to improve the low-light scene generalization ability. LLFormer~\cite{wang2023ultra} utilized a transformer-based approach with axis-based multi-head self-attention and cross-layer attention fusion block to effectively enhance low-light images. 
Although the aforementioned methods perform well in the LLE task, their direct concatenation with normal-light SR techniques for the LLISR task shows evident limitations.

\begin{figure*}[htbp]
	\centering 
	\begin{tabular}{c}
		\includegraphics[width=16cm]{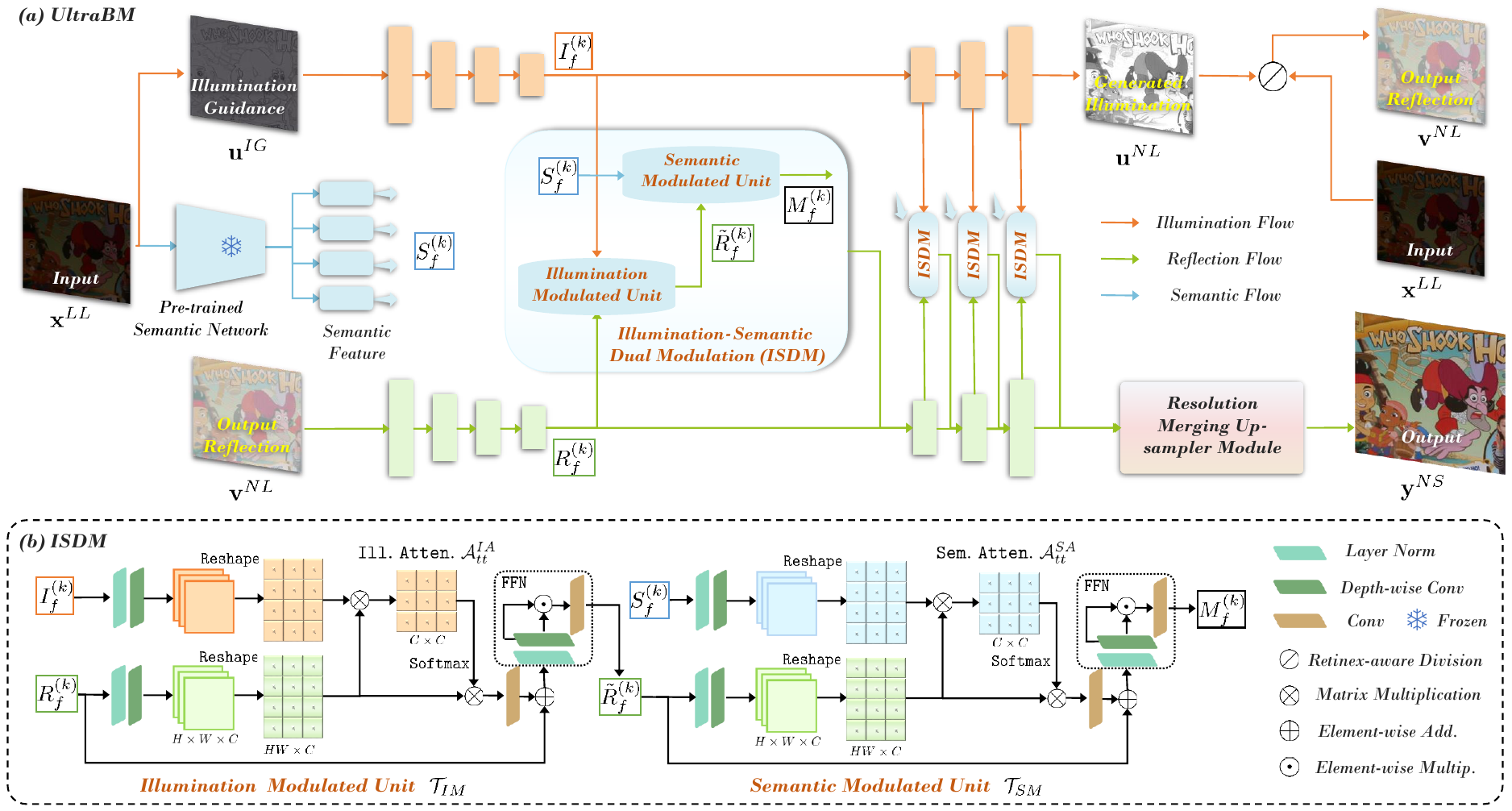}\\
	\end{tabular}%
	\caption{\textbf{\textit{(a)}}: The overview of the proposed UltraBM. Given low-light low-resolution image $\mathbf{x}^{LL}$, we construct the dual-stream framework to generate the initial reflection image $\mathbf{v}^{NL}$ with illumination guidance $\mathbf{u}^{IG}$. Subsequently, $\mathbf{v}^{NL}$ is fed into a subsequent refinement super-resolution network to restore a normal-light, super-resolution image $\mathbf{y}^{NS}.$ \textbf{\textit{(b)}} shows the refinement middleware, ISDM, which top-down modulates reflection features across branches to facilitate interactions.
	} \label{fig:workflow}
	\vspace{-3mm}
\end{figure*}%

\subsection{Normal-light Image Super-Resolution}  
Normal-light SR task involves generating high-resolution images from low-resolution inputs under standard lighting conditions. In recent decades, a large number of methods based on convolutional neural networks have emerged to continuously refresh the performance \cite{zhang2018residual,li2019feedback,zhao2020efficient,haris2018deep,wang2018esrgan}. 
It is worth noting that Zamir \textit{et al.}~\cite{zamir2020learning} proposed the image restoration architecture with a non-local attention mechanism and multi-scale feature aggregation, which demonstrates remarkable performance in normal-light SR tasks. 
Recently, benefiting from the surge of self-attention mechanisms, various transformer-based approaches have been developed for super-resolution enhancement. Among these, notable approaches include SwinIR~\cite{liang2021swinir}, Restormer~\cite{zamir2022restormer}, the recently proposed SRFormer~\cite{zhou2023srformer} and HAT~\cite{chen2023hat}.
However, these methods tailored for normal lighting conditions fail to address the challenges posed by low-light environments, often resulting in undesirable outcomes such as artifact spreading and texture blurring.

\section{Methodology} 
%
%

Focusing on LLISR, this paper aims to establish an end-to-end learning framework for recovering a normal-light super-resolution image (i.e., $\mathbf{y}^{NS}$) from a given low-resolution image captured in low-light environments (i.e., $\mathbf{x}^{LL}$). The crux of this approach lies in simultaneously learning brightness enhancement while ensuring an elevation in resolution. 
To tackle this issue effectively,  we introduce a tailored low-light super-resolution framework for LLISR, dubbed UltraBM. 
In what follows, we detail the proposed method. 

\subsection{Retinex-Inspired Dual-stream Framework}
Drawing inspiration from the classical Retinex theory, we initially construct a foundational illumination learning network to capture the underlying physical principles of low-light scenes. As illustrated in \textit{Figure~\ref{fig:workflow}} \textit{(a)}, we introduce a neighborhood difference operator $F_{diff}$ following~\cite{ma2021learning} to compute the initial illumination guidance $\mathbf{u}^{IG}=F_{diff}\big(\mathbf{x}^{LL}\big)$. Subsequently, $\mathbf{u}^{IG}$ is fed into a U-Net-style network to generate the illumination map $\mathbf{u}^{NL}$, with such learning process being supervised by the introduced scene constraint for untra-dark environments\footnote{Please refer to constraint losses as outlined in Eqs.~\eqref{eq:sl}-\eqref{eq:is}.}. Further, we obtain the low-resolution reflection map $\mathbf{v}^{NL}$ by employing the Retinex-based element-wise division, formulated as 
\begin{equation}
\mathbf{v}^{NL}=\mathbf{x}^{LL} \oslash \mathbf{u}^{NL},~\mathbf{u}^{NL}=Unet(\mathbf{u}^{IG}).
\end{equation}

\begin{figure}[!t]
	\centering 
	\begin{tabular}{c}
		\includegraphics[width=8.5cm]{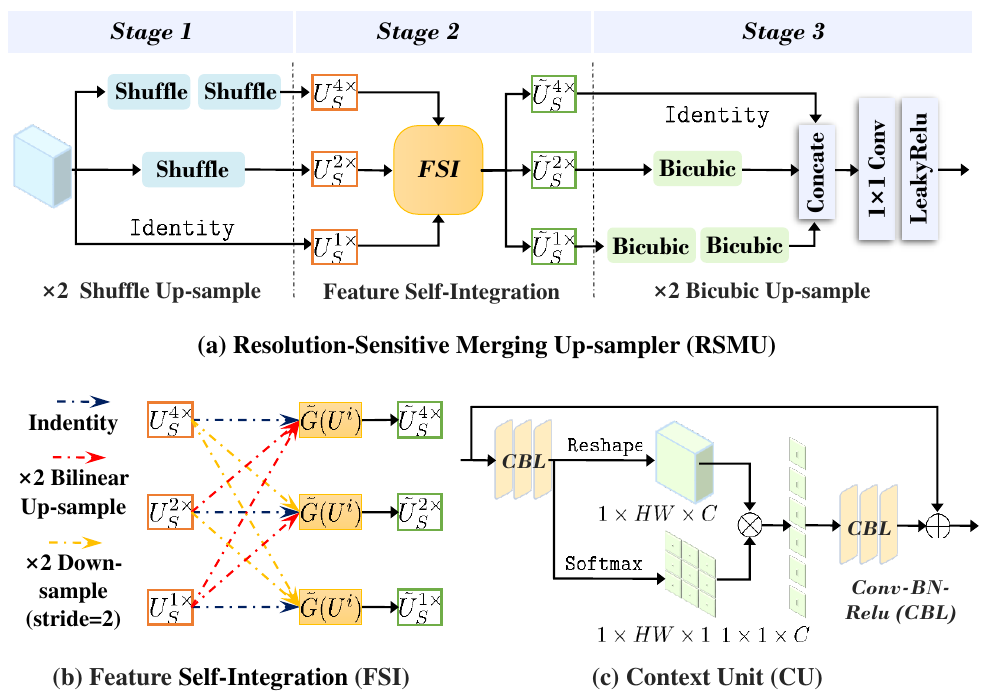}\\
	\end{tabular}%
	\caption{Illustrations of RSMU \textit{(a)}, FSI \textit{(b)} and CU \textit{(c)}. } \label{fig:RSMU}
	\vspace{-3mm}
\end{figure}

In the subsequent stage,  $\mathbf{v}^{NL}$ is transmitted to another Unet-style network with similar architecture to complete refined super-resolution process.   Inspired by self-attention mechanisms~\cite{zamir2022restormer}, we design a refined intermediary, namely Illumination-Semantic Dual Modulation (ISDM), depicted as the blue shaded box in \textit{Figure~\ref{fig:workflow}} \textit{(b)}, positioned within the decoding stage to facilitate progressive inter-branch interactions. 
Experiments confirm that ISDM can effectively avoid exacerbating color shifts and distortions during image super-resolution. 
Finally, we devised a Resolution-Sensitive Merging Up-sampler (RSMU) module  to obtain the enhanced image $\mathbf{y}^{NS}$. 
It is worth noting that the Unet is constructed based on Context Units (CUs, see \textit{Figure~\ref{fig:RSMU} (c)}) as the base module, which transforms the scale using Maxpooling and bilinear up-sampling operations in the encoding and decoding phases, respectively\footnote{Refer to the Supplementary Material for detailed architectures. }.

In the following, we expand on the architectural details of two key modules designed, including ISDM and RSMU.

\subsection{Illumination-Semantic Dual Modulation}

As shown in \textit{(a)} of \textit{Figure~\ref{fig:workflow}}, the ISDM consists of the Illumination Modulated Unit (IMU, $\mathcal{T}_{IM}$)  and Semantic Modulated Unit  (SMU, $\mathcal{T}_{SM}$)  with a similar architectural design.  The illumination features and reflection features extracted from the decoders of the top and down branches are denoted as $I^{(k)}_{f}$ and $R^{(k)}_{f}$, respectively. $k\in\{1,2,\cdots,5\}$ indexes the feature level.  Inspired by the fact that large semantic knowledge base can improve the network's representational capabilities, we utilize a pre-trained semantic network HRNet~\cite{wang2020deep} to extract multi-scale semantic features $S^{(k)}_{f}$ from $\mathbf{x}^{LL}$.

The ISDM facilitates a top-down information flow of illumination and semantic features, and calculates a similarity matrix as modulation response to guide the refinement of reflection features. The structural details of ISDM are depicted in \textit{Figure~\ref{fig:workflow}} \textit{(b)}.
Specifically, $I^{(k)}_{f}$ and $R^{(k)}_{f}$ are individually transformed through layer normalization and depth-wise convolution, further reshaping to compute the illumination attention map  $\mathcal{A}_{tt}^{IA}\in\mathcal{R}^{C \times C}$, presented by
\begin{equation} 
	\mathcal{A}_{tt}^{IA}~=~LN~\big(Conv~(I^{(k)}_{f})\big)~\otimes~LN~\big(Conv~(R^{(k)}_{f})\big).
\end{equation}
Then $\mathcal{A}_{tt}^{IA}$ is normalized through the softmax function and update $R^{(k)}_{f}$ in a dynamic weighting manner, and a feed-forward network $\psi_{FFN}$ followed by~\cite{zamir2022restormer} is introduced for further content reconstruction, expressed as 
\begin{equation} 
	\mathcal{T}_{IM}=\psi_{FFN}~\big(\mathcal{S}~[\mathcal{A}_{tt}^{IA}~\otimes~ {Conv}~(R^{(k)}_{f})]\big), 
\end{equation}
where $\mathcal{S}$ is the Sigmoid function, and $\otimes$ denotes the element-wise multiplication operation. 
 Thus we obtain the modulated reflection features  $\tilde{R}^{(k)}_{f}=\mathcal{T}_{IM}(I^{(k)}_{f},R^{(k)}_{f})$. 
Similarly, $\tilde{R}^{(k)}_{f}$ is fed into the proposed $\mathcal{M}^{SMU}$ to generate the semantic attention map $\mathcal{A}_{tt}^{SA}\in\mathcal{R}^{C \times C}$, followed by 
\begin{equation} 
	\mathcal{T}_{SM}=\psi_{FFN}~\big(\mathcal{S}~[\mathcal{A}_{tt}^{SA}~\otimes~ {Conv}~(\tilde{R}^{(k)}_{f})]\big), 
\end{equation}
where $\mathcal{A}_{tt}^{SA}~=~LN~\big(Conv~(S^{(k)}_{f})\big)~\otimes~LN~\big(Conv~(\tilde{R}^{(k)}_{f})\big)$. 
The doubly modulated feature $M^{(k)}_{f}=\mathcal{T}_{SM}(S^{(k)}_{f},\tilde{R}^{(k)}_{f})$ is then propagated to the  $k+1$ decoding layer.

\subsection{Resolution-Sensitive Merging Up-sampler}
Typically, most of the normal-light SR  methods employ naive up-sampling techniques (e.g., Bilinear~\cite{zhang2018image} or Bicubic), which lack the ability to fully exploit the collaborative potential of scale-sensitive features. As a consequence, these methods face challenges in reproducing intricate high-quality details, undesirable artifacts and halos. Motivated by this, we construct the RSMU module that amalgamates distinct sampling modalities as foundational constituents to progressively learn the mapping from low-resolution space to high-resolution space. 

As illustrated in \textit{Figure~\ref{fig:RSMU}} \textit{(a)}, RSMU comprises three stages, each employing different sampling modalities as substrates, including pixel shuffle up-sampling, bilinear up-sampling, and bicubic up-sampling.  Initially, it employs pixel shuffle in parallel to generate three separate scale features $U_S^{i\times}$, where $i \in \{1, 2, 4\}$ indexes the scale layer. 
Subsequently, these three sets of distinct features are converged within the Feature Self-Integration (FSI) module for both feature selection and fusion.  
As depicted in \textit{Figure~\ref{fig:RSMU}} \textit{(b)}, FSI aggregates and transforms the input features $U_S^{i\times}$, yielding corresponding outputs denoted as $\tilde{U}_S^{j\times}$,  where $j$ takes values from  $\{1, 2, 4\}$ to represent different output resolution levels.  Through the introduced selective attention mechanism \cite{zamir2022learning}, denoted as $\mathcal{M}_{skff}$, each $\tilde{U}_S^{j\times}$ dynamically selects essential features, which can be formalized as follows:
\begin{equation} \small
	\tilde{U}_S^{j\times}=\lfloor\tilde{G}(U_S^{i\times})\rfloor_{i=1,2,4},\  \tilde{G}=\mathcal{M}_{skff}\circ\left\{\begin{array}{ccc}1&\forall&i=j,\\\frac{j}{i}\uparrow&\forall&j>i,\\\frac{i}{j}\downarrow&\forall&j<i.\end{array}\right.
\end{equation}
In the specific mathematical form of $\tilde{G}$: 
when $i=j$, an Identity operation (with unchanged size) is used (blue dashed line); 
when $j<i$, a transposed convolution down-sampling with a stride of 2 is employed (yellow dashed line); 
when $j>i$, bilinear up-sampling is used (red dashed line).

\begin{figure*}[!h]
	\centering 
	\begin{tabular}{c}
		\includegraphics[width=17.5cm]{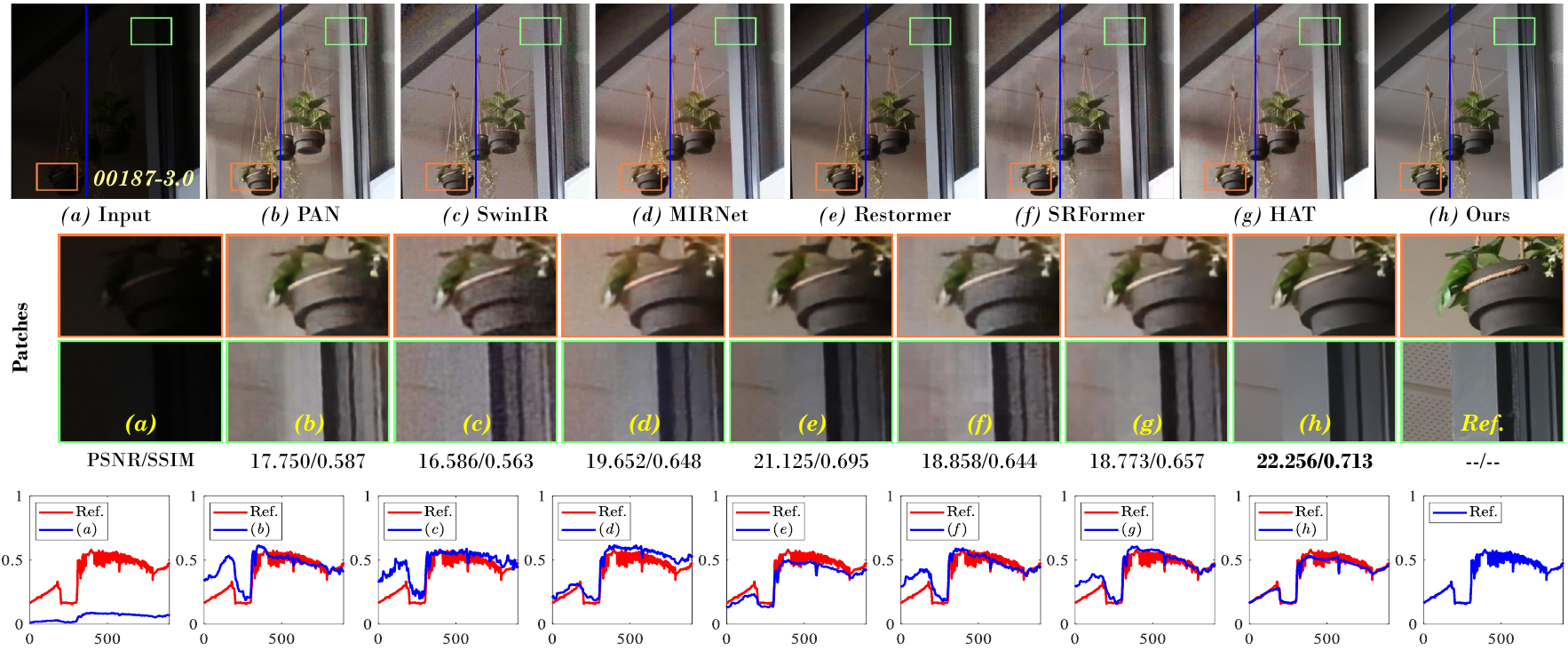}\\
	\end{tabular}%
	\caption{Qualitative comparisons on RELLISUR dataset of our method with SOTA methods for 2x LLISR task.  Below signal maps provide the differences of pixel intensity between generated images and the reference image along the blue line segment.} \label{fig:Res1}
\end{figure*}%


\begin{table*}[!h]
	\small
	\centering
	\tabcolsep=2pt
	\renewcommand{\arraystretch}{1.2}
	\scalebox{0.84}{	
		\begin{tabular}{cccccccccc}
			\toprule[1.5pt] 
			\multirow{1}{*}{\emph{\textbf{Metrics}}} &  RDN$^{18CVPR}$ & SRFBN$^{19CVPR}$ & PAN$^{20ECCV}$ & MIRNet$^{20ECCV}$ & SwinIR$^{21ICCV}$  & Restormer$^{22CVPR}$ & SRFormer$^{23ICCV}$ & HAT$^{23CVPR}$ & Ours \\			
			\hline
			\rule{0pt}{11pt}
			PSNR$\uparrow$ & 18.794/18.201 &18.435/17.653&18.757/18.051&21.054/19.773&18.387/17.452&\underline{\textbf{21.217}}/\underline{\textbf{20.290}}&19.558/18.708&20.213/19.750&\textcolor{black}{\textbf{22.264}}/\textcolor{black}{\textbf{21.036}}\\	
			\cdashline{2-10}[2pt/5pt]
			\rule{0pt}{11pt}
			SSIM$\uparrow$ & 0.701/0.701&0.662/0.665&0.716/0.703 &0.720/0.704& 0.647/0.658&\underline{\textbf{0.727}}/\underline{\textbf{0.720}}&0.704/0.705&0.719/0.715&\textcolor{black}{\textbf{0.743}}/\textcolor{black}{\textbf{0.726}}\\
			\cdashline{2-10}[2pt/5pt]
			\rule{0pt}{11pt}
			LPIPS$\downarrow$&0.455/0.584 &0.510/0.640&0.465/0.567 &0.436/0.599&0.577/0.686&\underline{\textbf{0.385}}/\underline{\textbf{0.492}}&0.469/0.613&0.454/0.561&\textcolor{black}{\textbf{0.254}}/\textcolor{black}{\textbf{0.371}}\\	
			\cdashline{2-10}[2pt/5pt]
			\rule{0pt}{11pt}
			RMSE$\downarrow$&0.120/0.128 &0.125/0.136&0.119/0.129 &\underline{\textbf{0.095}}/0.109&0.125/0.139&\underline{\textbf{0.095}}/\underline{\textbf{0.106}}&0.110/0.430&0.103/0.110&\textcolor{black}{\textbf{0.005}}/\textcolor{black}{\textbf{0.006}}\\				
			\hline
			\rule{0pt}{11pt}
			NIQE$\downarrow$ &7.864/9.098&\textcolor{black}{\textbf{7.202}}/\underline{\textbf{7.766}}&7.370/8.737&7.881/8.803&7.231/9.489&7.636/9.065&7.550/9.664&8.332/10.028&\underline{\textbf{7.216}}/\textcolor{black}{\textbf{7.537}}\\
			\cdashline{2-10}[2pt/5pt]
			\rule{0pt}{11pt}
			LOE$\downarrow$ &45.642/44.976&47.698/45.582&49.793/48.623&31.968/33.966&46.042/42.581 &\underline{\textbf{30.889}}/\underline{\textbf{30.856}}&50.411/45.606&42.265/35.284&\textcolor{black}{\textbf{28.670}}/\textcolor{black}{\textbf{29.432}}\\		
			\bottomrule[1.5pt] 
		\end{tabular}
	}
\caption{Quantitative comparison among various normal-light SR methods on \textit{RELLISUR-Test} dataset (i.e.,2x / 4x).  The top-ranked and the second-ranked method are highlighted in bold and bold underlined, respectively. } 
	\label{tab:rellisur2}
\end{table*}

\subsection{Loss Function}
\noindent\textit{\textbf{Self-regularized Luminance Loss.}} Inspired by the color statistical regularities of natural image distributions, we propose the self-regularized luminance loss $\mathcal{L}_{SL}$, to encourage the naturalness of colors for $\mathbf{v}^{NL}$, which can be formulated as 
\begin{equation} 
	\mathcal{L}_{SL}(\mathbf{v}^{NL})=e^{|\bar{\mathbf{v}}_c^{NL}-\bm{\mu}_c-\bm{\sigma}_c|}-1, \ c\in\{R,G,B\}, \label{eq:sl}
\end{equation}
where $\bm{\mu}_c$ and $\bm{\sigma}_c$ denote mean and standard deviation of the natural image distribution of Imagenet~\cite{deng2009imagenet}, with $\bm{\mu}_c=[0.485,0.456,0.406]$ and $\bm{\sigma}_c=[0.229,0.224,0.225]$. $\bar{\mathbf{v}}^{NL}$ denotes the channel-wise mean. 

\noindent\textit{\textbf{Illumination Smooth Loss.}} Drawing insights from~\cite{ma2021learning},  we utilize the  $\mathtt{Smooth}_{\mathcal{L}_1}$ loss to ensure the structural consistency between the generated illuminance image $\mathbf{u}^{NL}$ and the gray image $\mathbf{x}_{G}^{LL}$ of input, presented as 
\begin{equation} 
	\mathcal{L}_{IS}(\mathbf{u}^{NL},\mathbf{x}_{G}^{LL})=\mathtt{Smooth}_{\mathcal{L}_1}\left(\mathbf{u}^{NL}-\mathbf{x}_{G}^{LL}\right),\label{eq:is}
\end{equation}
%
where $\mathtt{Smooth}_{\mathcal{L}_1}(x)=0.5x^2$ when $|x|\leq 1$; otherwise, $\mathtt{Smooth}_{\mathcal{L}_1}(x)=|x|-0.5$. 

\noindent\textit{Reconstruction Loss.} To maintain the content consistency between the generated image $\mathbf{y}^{NS}$ and the reference image (normal-light high-resolution image) $\mathbf{y}^{NH}$, we introduce the reconstruction loss, i.e., 
\begin{equation}
	\mathcal{L}_{R} (\mathbf{y}^{NS},\mathbf{y}^{NH}) = ||\mathbf{y}^{NS}-\mathbf{y}^{NH}||_1, 
\end{equation}

\noindent\textit{\textbf{Perceptual Loss.}}
To maintain perceptual consistency, we introduce perceptual loss~\cite{johnson2016perceptual}  to calculate the disparity between $\mathbf{y}^{NS}$ and $\mathbf{y}^{NH}$:
\begin{equation}
	\mathcal{L}_{P} (\mathbf{y}^{NS},\mathbf{y}^{NH}) = \frac{1}{c_j h_j w_j}||\phi_j(\mathbf{y}^{NS})-\phi_j(\mathbf{y}^{NH})||_1, 
\end{equation}
where $\phi$ is the VGG-19 model, $c_j h_j w_j$ denotes the size of the feature map at the j-th layer. We use $\{\mathtt{conv1},\cdots, \mathtt{conv5}\}$ feature layers with weights $\{0.1,0.1,1,1,1\}$.  

\noindent\textit{\textbf{Overall Loss.}} We train our network by minimizing the following overall loss: 
\begin{equation}\label{eq:loss}
	\mathcal{L}_{total}={\lambda}_{1}*\mathcal{L}_{SL}+{\lambda}_{2}*\mathcal{L}_{IS}+{\lambda}_{3}*\mathcal{L}_{R}+{\lambda}_{4}*\mathcal{L}_{P},
\end{equation}
where the weights $\{{\lambda}_{i}\}_{i=1}^4$ are set to 1.0, 1.0, 1.0, 1.2. 

\begin{table}[!h]
	\scalebox{0.9}{	
		\begin{threeparttable}   
			\small
			\centering
			\tabcolsep=2pt
			\renewcommand{\arraystretch}{1.2}	
			\begin{tabular}{cccc}
				\toprule[1.5pt] 			 
				\multirow{2}{*}{Metrics} & ZeroDCE$^{21TPAMI}_{\dagger}$ & SCI$^{22CVPR}_{\dagger}$ & LLFormer$^{23AAAI}_{\dagger}$ \\
				&$\Rightarrow$HAT$^{23CVPR}_{\ddagger}$  &$\Rightarrow$HAT$^{23CVPR}_{\ddagger}$ & $\Rightarrow$HAT$^{23CVPR}_{\ddagger}$ \\				
				\hline
				\rule{0pt}{11pt}
				PSNR$\uparrow$ &12.927/12.524 & 14.963/14.776 & 21.286/20.135  \\	
				\cdashline{2-4}[2pt/5pt]
				\rule{0pt}{11pt}
				SSIM$\uparrow$ &0.354/0.321& 0.439/0.452& 0.720/0.718 \\
				\cdashline{2-3}[2pt/5pt]
				\rule{0pt}{11pt}
				LPIPS$\downarrow$&0.698/0.739& 0.591/0.697 & 0.455/0.575\\	
				\cdashline{2-4}[2pt/5pt]
				\rule{0pt}{11pt}
				RMSE$\downarrow$&0.015/0.019 & 0.200/0.205 & 0.093/0.105\\	
				\cdashline{1-4}[2pt/5pt] 
				\rule{0pt}{11pt}
				NIQE$\downarrow$&7.389/8.132 & 7.625/8.235  & 8.574/9.231 \\	
				\cdashline{2-4}[2pt/5pt] 
				\rule{0pt}{11pt}
				LOE$\downarrow$&56.726/54.178  &  36.413/37.879  & 35.084/36.215 \\	
				\bottomrule[1.5pt] 
			\end{tabular}
			\begin{tablenotes}[para,flushleft]   
				\item  $\dagger$ indicates training on 1x low-light RELLISUR dataset for LLE. $\ddagger$ denotes training on  2x or 4x \textit{RELLISUR} dataset for normal-light SR. 
			\end{tablenotes} 				
			\caption{Quantitative comparison among cascaded LLE$\Rightarrow$SR methods on \textit{RELLISUR} dataset (i.e.,2x / 4x). } 	
			\label{tab:rellisur3} 
		\end{threeparttable}
	}
\end{table} 

\section{Experiments}  


\subsection{Dataset and Experimental Settings}  

\noindent\textit{\textbf{Dataset.}} 
We evaluate the performance on two widely-used datasets: RELLISUR\footnote{https://vap.aau.dk/rellisur/} and DarkFace~\cite{poor_visibility_benchmark}\footnote{https://flyywh.github.io/CVPRW2019LowLight/}. We utilize the RELLISUR dataset for training our method, since RELLISUR offers a collection of 1,045 image pairs at three distinct resolution scales (1x, 2x, and 4x) and five varying low-light levels (ranging from -2.5EV to -5.0EV), encompassing both low-resolution low-light images and high-resolution normal-light images.

\noindent\textit{\textbf{Evaluation Metrics.}}
We evaluate the performance using four widely-used full-reference metrics, including pixel-wise Peak Signal to Noise Ratio (PSNR), Structural Similarity Index (SSIM), \textit{LPIPS}~\cite{agaian2007transform} and Root Mean Square Error (RMSE), as well as two no-reference metrics Natural Image Quality Evaluator (NIQE)~\cite{mittal2012making} and LOE~\cite{wang2013naturalness}.


\noindent\textit{\textbf{Implement Details.}} 
All experiments were performed on a PC equipped with an NVIDIA GeForce GTX 2080Ti GPU, using the PyTorch 1.8.0 framework. Our model was trained using the Adam optimizer for a total of 150,000 iterations. The initial learning rate was set to $2 \times 10^{-4}$ and the weight decay parameter is $1 \times 10^{-4}$ with $\beta=[0.9, 0.999]$.  The progressive training strategy with mini batch sizes sets to $[8,5,4,2,1,1]$. 
\begin{figure*}[!h]
	\centering 
	\begin{tabular}{c}
		\includegraphics[width=17.5cm]{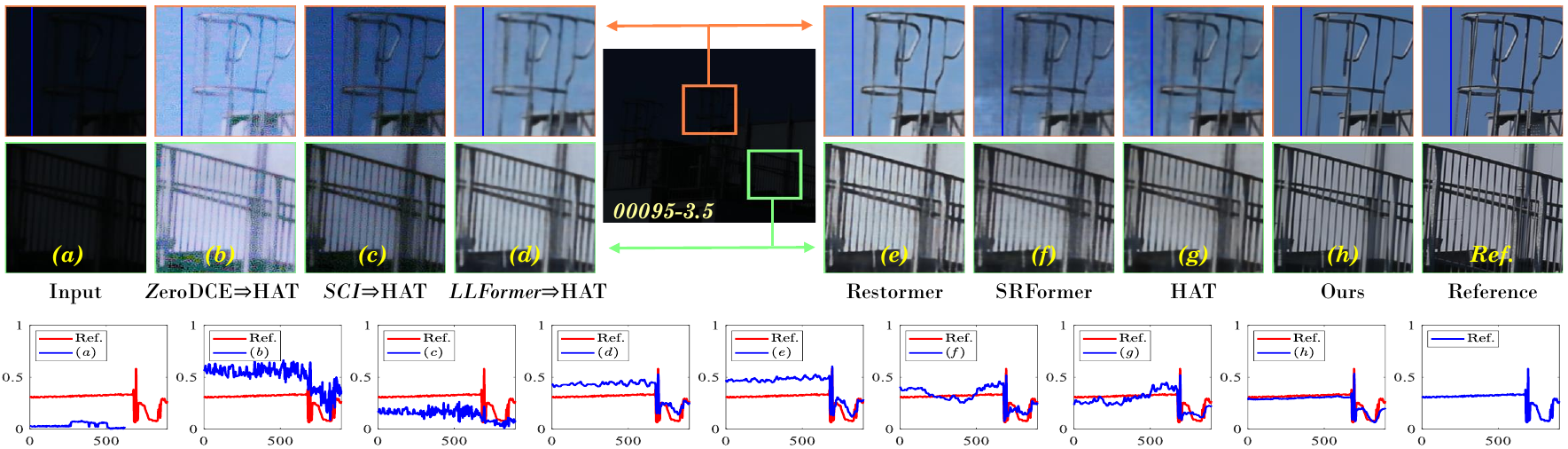}\\
	\end{tabular}%
	\caption{Qualitative comparisons on RELLISUR dataset for 4x LLISR task. Below provides the differences of pixel intensity.} \label{fig:Res2}
\end{figure*}%
\begin{figure*}[htbp]
	\centering 
	\begin{tabular}{c}
		\includegraphics[width=17.5cm]{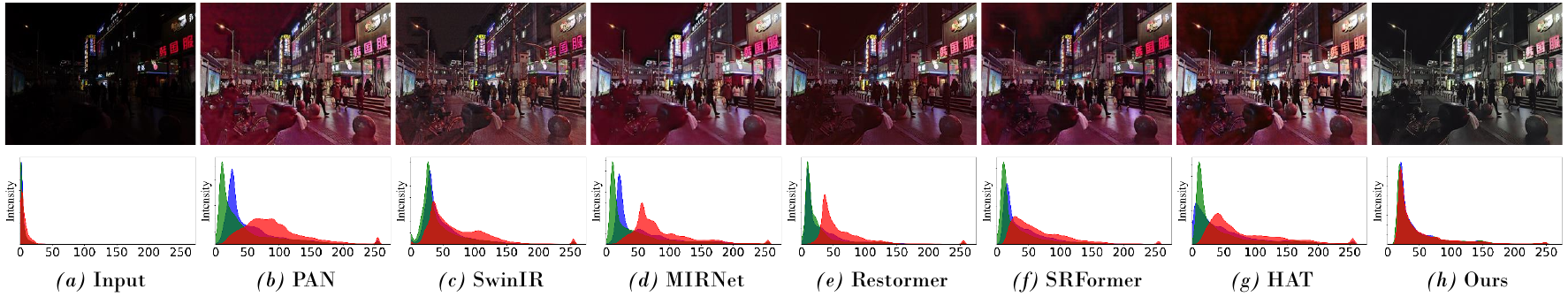}\\
	\end{tabular}%
	\caption{Visual comparison on DarkFace dataset for 2x LLISR task. Below is the probability density histogram about RGB. } \label{fig:darkface2}
\end{figure*}%

\subsection{Comparisons with State-of-the-Art}
We compare our method against two distinct schemes: normal-light SR methods and cascaded LLE$\Rightarrow$SR methods. 
To ensure a comprehensive comparison, we meticulously selected three representative LLE methods, including ZeroDCE~\cite{li2021learning}, SCI~\cite{ma2022toward}, and LLFormer~\cite{wang2023ultra}, and eight normal-light SR methods, i.e., RDN~\cite{zhang2018residual}, SRFBN~\cite{li2019feedback}, PAN~\cite{zhao2020efficient},   SwinIR~\cite{liang2021swinir}, and MIRNet~\cite{zamir2020learning}, Restormer~\cite{zamir2022restormer}, SRFormer~\cite{zhou2023srformer}, HAT~\cite{chen2023hat}. 
We retrain all methods on RELLSUR for fair comparisons.

\begin{figure}[!h]
	\centering 
	\begin{tabular}{l}
		\includegraphics[width=8.2cm]{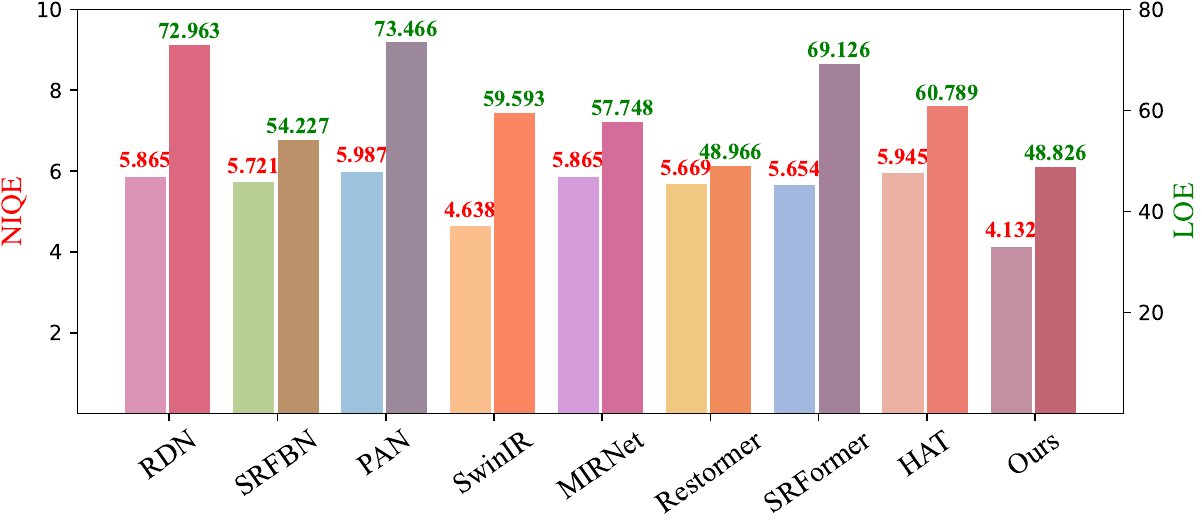}\\
	\end{tabular}%
	\caption{Generalization analysis on \textit{DarkFace} dataset.} \label{fig:darkface}
\end{figure}%

\begin{figure}[!h]
	\centering 
	\begin{tabular}{c}
		\includegraphics[width=8.2cm]{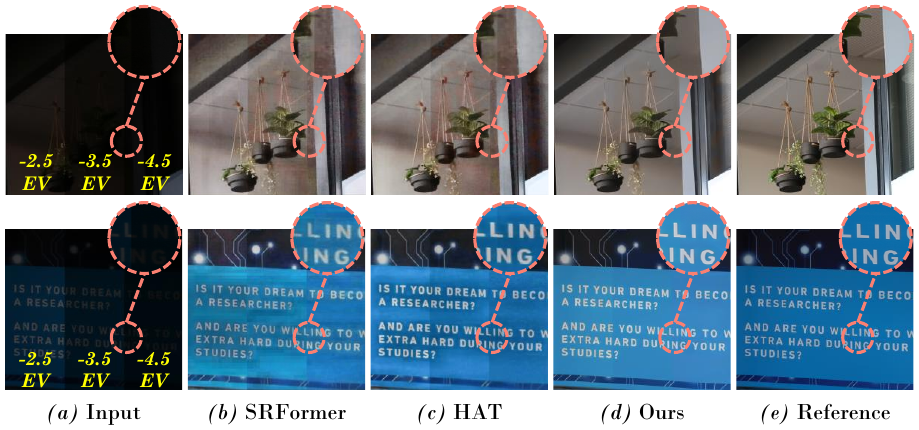}\\
	\end{tabular}%
	\caption{Generalization analysis across diverse darkness levels (ranging from -2.5EV to -4.5EV).} \label{fig:darkface3}
\end{figure}%

\begin{table}[!h]
	\small
	\centering
	\tabcolsep=2pt
	\renewcommand{\arraystretch}{1.2}
	\scalebox{0.70}{	
		\begin{tabular}{cccccc}
			\toprule[1.5pt] 
			\multirow{1}{*}{\emph{\textbf{Metrics}}} & SwinIR  & Restormer & SRFormer  & HAT  & Ours \\			
			\hline
			\rule{0pt}{11pt}
			Params (MB) &11.683/11.825&26.126/26.209 &\underline{\textbf{10.162}}/\underline{\textbf{10.220}}&\textcolor{black}{\textbf{9.473/9.621}}&69.561/69.687 \\
			\cdashline{2-3}[2pt/5pt]
			\rule{0pt}{11pt} 
			FLOPs (G) &453.12/467.94&\underline{\textbf{35.375}}/\underline{\textbf{41.327}}&81.797/83.006&58.990/68.994&\textcolor{black}{\textbf{31.497/64.633}} \\
			\cdashline{2-3}[2pt/5pt]
			\rule{0pt}{11pt} 
			Infer. (S) &3.879/4.226&\textcolor{black}{\textbf{0.033/0.035}}&0.218/0.221 &0.184/0.186&\underline{\textbf{0.044}}/\underline{\textbf{0.054}} \\		
			\bottomrule[1.5pt] 
		\end{tabular}
	}	
	\caption{Computational efficiency of SOTA methods (2x/4x).} \label{tab:CE}
\end{table}

\noindent\textit{\textbf{LLE$\Rightarrow$SR Methods.}} Among numerous SR methods, we have chosen the latest SR method, HAT$^{23CVPR}$, as the subsequent magnification model, cascaded behind three LLE methods. This setup allows us to observe the performance of the concatenated methods for the LLISR problem. Notably, considering that ZeroDCE and SCI are unsupervised methods, we solely trained them on the 1x low-light RELLISUR dataset for the preliminary brightening task. Additionally, as HAT serves as a posterior module, used for super-resolution on the brightened images, we trained the model on the 2x and 4x normal-light RELLISUR datasets, denoting it as HAT$_{\ddagger}$ to distinguish it from the original HAT.

\begin{table}[htb]	
	\small
	\centering
	\renewcommand{\arraystretch}{1.2}
	\scalebox{0.95}{	
		\begin{tabular}{lccc}
			\toprule[1.5pt]
			\rule{0pt}{8pt} 
			\multirow{1}{*}{\textbf{Configuration}} 
			& PSNR$\uparrow$ & SSIM$\uparrow$ & LPIPS$\downarrow$  \\
			\hline
			\rule{0pt}{11pt}
			OS & 20.213$\textcolor{black}{_{\downarrow\textbf{2.051}}}$ & 0.676$\textcolor{black}{_{\downarrow\textbf{0.067}}}$ & 0.316$\textcolor{black}{_{\uparrow\textbf{0.062}}}$\\
			\cdashline{2-4}[2pt/5pt]
			\rule{0pt}{11pt}
			DS (w/ $\mathbf{x}^{LL}$) & 21.821$\textcolor{black}{_{\downarrow\textbf{0.443}}}$ & 0.713$\textcolor{black}{_{\downarrow\textbf{0.030}}}$& 0.285$\textcolor{black}{_{\uparrow\textbf{0.031}}}$ \\
			\cdashline{2-4}[2pt/5pt]
			\rule{0pt}{11pt}
			DS (w/ $\mathbf{v}^{LL}$)  &\textcolor{black}{\textbf{22.264}} &	\textcolor{black}{\textbf{0.743}} &	\textcolor{black}{\textbf{0.254}}  \\ 
			\bottomrule[1.5pt] 			
	\end{tabular} } 
	\caption{Exploring the performance comparison between one-stream (OS) and dual-stream (DS) architectures. 
	} \label{tab:os}
\end{table}

\noindent\textit{\textbf{Quantitative Evaluation.}}  
\textit{Tables}~\ref{tab:rellisur2} and \ref{tab:rellisur3} present the quantitative results among various normal-light SR methods and cascaded methods on the \textit{RELLISUR} dataset for 2x and 4x tasks. Our method outperforms existing state-of-the-art  (SOTA) approaches, achieving the highest scores in six evaluation metrics. 
Compared to the closely ranked second-best Restormer, our approach achieved a significant improvement (i.e., $\uparrow$5\%  in PSNR, $\uparrow$2\%  in SSIM, and $\uparrow$50\% in LPIPS). 

\noindent\textit{\textbf{Qualitative Evaluation.}}  
Qualitative results on realistic \textit{RELLISUR} dataset  are displayed in \textit{Figures}~\ref{fig:Res1} and~\ref{fig:Res2}. As illustrated, our method compared to other methods produces authentic images with vivid lightness, and an exceptional ability to recover high-frequency structural details. 
\textit{Figure}~\ref{fig:darkface2} displays visualization results on the \textit{DarkFace-Test} dataset. 
Particularly noteworthy is our method's exceptional performance in maintaining the natural authenticity of nighttime scenes. 

\begin{table}[!h]
	\scalebox{0.95}{		
		\small
		\centering
		\renewcommand{\arraystretch}{1.2}
		\begin{tabular}{lccc}
			\toprule[1.5pt]
			\rule{0pt}{8pt} 
			\multirow{1}{*}{\textbf{Configuration}} 
			& PSNR$\uparrow$ & SSIM$\uparrow$ & LPIPS$\downarrow$  \\
			\hline
			\rule{0pt}{11pt}
			\textit{(a)} SBMNet$^{\bm{\dagger}}$ &\textcolor{black}{\textbf{21.954}} &	\textcolor{black}{\textbf{0.735}} &	\textcolor{black}{\textbf{0.268}}  \\ 
			\cdashline{2-4}[2pt/5pt]
			\rule{0pt}{11pt}
			\textit{(b)} w/o ISDM & 21.401$\textcolor{black}{_{\downarrow\textbf{0.553}}}$ & 	0.726$\textcolor{black}{_{\downarrow\textbf{0.009}}}$ &	0.285$\textcolor{black}{_{\uparrow\textbf{0.017}}}$   \\
			\cdashline{2-4}[2pt/5pt]
			\rule{0pt}{11pt}
			\textit{(c)} ISDM - SMU & 21.635$\textcolor{black}{_{\downarrow\textbf{0.319}}}$ & 	0.728$\textcolor{black}{_{\downarrow\textbf{0.007}}}$ &	0.279$\textcolor{black}{_{\uparrow\textbf{0.011}}}$  \\
			\cdashline{2-4}[2pt/5pt]
			\rule{0pt}{11pt}
			\textit{(d)} ISDM - IMU & 21.732$\textcolor{black}{_{\downarrow\textbf{0.222}}}$ & 	0.730$\textcolor{black}{_{\downarrow\textbf{0.005}}}$ &	0.281$\textcolor{black}{_{\uparrow\textbf{0.013}}}$  \\	
			\cdashline{1-4}[2pt/5pt]
			\rule{0pt}{11pt}
			\textit{(e)} Bilinear& 21.550$\textcolor{black}{_{\downarrow\textbf{0.404}}}$ & 0.725$\textcolor{black}{_{\downarrow\textbf{0.010}}}$& 0.277$\textcolor{black}{_{\uparrow\textbf{0.009}}}$ \\
			\cdashline{2-4}[2pt/5pt]
			\rule{0pt}{11pt}
			\textit{(f)} RSMU - FSI & 21.653$\textcolor{black}{_{\downarrow\textbf{0.301}}}$ & 0.731$\textcolor{black}{_{\downarrow\textbf{0.004}}}$ & 0.275$\textcolor{black}{_{\uparrow\textbf{0.007}}}$\\
			\bottomrule[1.5pt] 			
		\end{tabular} 
	} 
	\caption{Ablation of the ISDM and RSMU. The subscript for models \textit{(b-f)} indicates the performance gap compared to the model \textit{(a)}. SBMNet$^{\bm{\dagger}}$ denotes training w/o the loss term $\mathcal{L}_{SL}$.} \label{tab:ISDM} 	
\end{table}


\noindent\textit{\textbf{Generalization Across Diverse Darkness Levels.}} \textit{Figure}~\ref{fig:darkface3} demonstrates the highly-robust generalization ability of our model across different levels of darkness. Its generalization is evident in maintaining excellent adaptability even under extremely short exposure conditions. In \textit{Tables}~\ref{tab:rellisur2} and \ref{tab:rellisur3}, the 19-fold higher RMSE score compared to other methods also underscores the remarkable generalization of our approach across various darkness levels in images.
\begin{table}[!h]
	
	\small
	\centering
	\renewcommand{\arraystretch}{1.2}
	\scalebox{0.85}{	
		\begin{tabular}{ccccccc}
			\toprule[1.5pt]
			\rule{0pt}{8pt} 
			
			$\mathcal{L}_{SL}$ & $\mathcal{L}_{IS}$ & $\mathcal{L}_{R}$ & $\mathcal{L}_{P}$ & PSNR$\uparrow$ & SSIM$\uparrow$ & LPIPS$\downarrow$  \\
			\hline
			\rule{0pt}{11pt}
			&\ding{51}& \ding{51} &\ding{51}&21.954$\textcolor{black}{_{\downarrow\textbf{0.310}}}$&0.735$\textcolor{black}{_{\downarrow\textbf{0.008}}}$&0.268$\textcolor{black}{_{\uparrow\textbf{0.014}}}$\\
			\cdashline{5-7}[2pt/5pt]
			\rule{0pt}{11pt}
			\ding{51}&\ding{51}&\ding{51}  &&22.149$\textcolor{black}{_{\downarrow\textbf{0.115}}}$&0.734$\textcolor{black}{_{\downarrow\textbf{0.009}}}$&0.417$\textcolor{black}{_{\uparrow\textbf{0.163}}}$\\
			\cdashline{5-7}[2pt/5pt]
			\rule{0pt}{11pt}
			\ding{51} &\ding{51}& \ding{51} &\ding{51}&\textbf{22.264}&\textbf{0.743}&\textbf{0.254}\\
			\bottomrule[1.5pt] 			
		\end{tabular} 
	} 
	\caption{Ablation of the different losses ($\mathcal{L}_{SL}$, $\mathcal{L}_{IS}$, and $\mathcal{L}_{P}$).} 
	\label{tab:loss}
\end{table}  

\noindent\textit{\textbf{Computational Efficiency.}} 
To examine model efficiency, we reported the \textit{Parameters(MB)}$\downarrow$, \textit{FLOPs(G)}$\downarrow$ and \textit{Inference(S)}$\downarrow$ of compared methods in \textit{Table}~\ref{tab:CE}. The measurements are conducted on a single 2080Titan GPU using images of size 128 $\times$ 128. Our method achieves a good balance between performance and computational efficiency. 

\subsection{Ablation Study} \label{sec:as}
\noindent\textit{\textbf{Why Design the Dual-stream Learning Framework?}} As depicted in \textit{Table}~\ref{tab:os}, removing the dual-stream branch and retaining only a single second-stage learning process results in a substantial 9\% reduction in the most severe PSNR score degradation. Additionally, substituting the input of the second stage with $\mathbf{x}^{LL}$ leads to a decrease of 0.443 dB in PSNR. These observations collectively underscore the significance of applying prior constraints to low-light scenes.

\begin{figure}[!t]
	\centering 
	\begin{tabular}{c}
		\includegraphics[width=8.5cm]{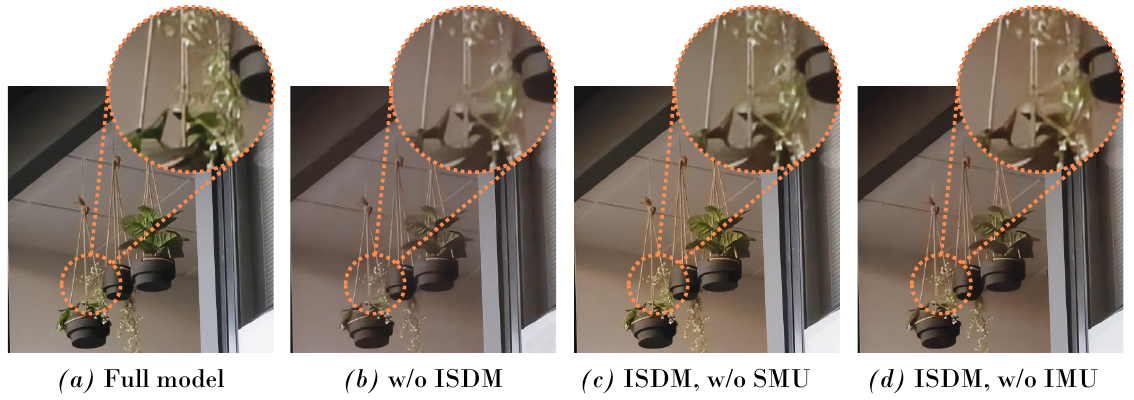}\\
	\end{tabular}%
	\caption{Ablation analysis with (w/) and without (w/o) the ISDM (also including the SMU, IMU components). } \label{fig:abl1}
\end{figure}

\noindent\textit{\textbf{Impact of ISDM.}} \textit{Table}~\ref{tab:ISDM} in [\textit{Config. \textit{(b)}}-\textit{Config. \textit{(d)}}]  and \textit{Figure}~\ref{fig:abl1} demonstrate the effectiveness of the ISDM module, including the efficacy of its intermediate components IMU and SMU. Compared to the baseline model [\textit{Config. \textit{(a)}}], removing the ISDM structure results in a significant performance drop (e.g., 0.553dB decrease in terms of PSNR). Similarly, removing its components (i.e., w/o SMU  [\textit{Config. \textit{(c)}}] or [\textit{Config. \textit{(d)}}]) also leads to a decrease in performance. 

\noindent\textit{\textbf{Effects of RSMU.}}  \textit{Table}~\ref{tab:ISDM} in [\textit{Config. \textit{(e)}}-\textit{Config. \textit{(f)}}] presents the effectiveness of RSMU. In comparison to  a simple up-sampling layer (e.g., Bilinear), RSMU facilitates a 0.404dB PSNR increase [\textit{Config. \textit{(e)}}]. The FSI module [\textit{Config. \textit{(f)}}] also contribute to the improvement in the final results. 

\noindent\textit{\textbf{Analysis of Different Losses.}} 
\textit{Table}~\ref{tab:loss} illustrates the ablation results for various loss components. We observe that the illuminance constraint (i.e., $\mathcal{L}_{SL}$) leads to an increase of 0.31 dB and 0.014 in PSNR and LPIPS scores, respectively. Additionally, the perceptual loss $\mathcal{L}_P$ has the most significant impact on the perceptual LPIPS scores.

\section{Conclusion and Remark}

This study addresses the relatively unexplored challenge of super-resolution in ultra-dark environments. Leveraging a novel dual-modulated learning framework, we introduced specialized components to enhance feature-level preservation of illumination and color details, while mitigating artifacts through a resolution-sensitive merging up-sampler. Our approach's remarkable applicability and generalizability across diverse ultra-low-light conditions were validated through comprehensive experiments. This work represents a significant advancement in addressing the complexities of super-resolution tasks under challenging low-light conditions. 

\textit{\textbf{Broader Impacts.}} Our work holds significance beyond the specific problem of ultra-dark super-resolution. This broader perspective reinvigorates the research landscape, encouraging exploration of joint multiple image processing tasks (i.e., low-light derain, low-light defog) in diverse adverse conditions. 
\bibliography{aaai24}

\end{document}